\documentclass[11pt]{article}

\usepackage{float} 
\usepackage{times}
\usepackage{epsfig}
\usepackage{graphicx}
\usepackage{amsmath}
\usepackage{amssymb}
\usepackage{bm}
\usepackage{algorithm}
\usepackage{algorithmicx}
\usepackage{algpseudocode}
\usepackage{comment}
\usepackage{fullpage}
\usepackage{tikz}
\usepackage{pifont}
\usepackage{amsthm}
\usepackage{subfigure}

\usepackage{authblk}
\newcommand{\myfootnote}[1]{
\renewcommand{\thefootnote}{}
\footnotetext{\hspace{-16.5pt}\scriptsize#1}
\renewcommand{\thefootnote}{\arabic{footnote}}}

\newcommand{\cmark}{\ding{51}}%
\newcommand{\xmark}{\ding{55}}

\newtheorem{lemma}{Lemma} 

\begin{document}

\title{Blind Hyperspectral-Multispectral Image Fusion via Graph Laplacian Regularization}
\author{Chandrajit Bajaj, Tianming Wang}
\affil{Institute for Computational Engineering and Sciences, University of Texas at Austin, Austin, Texas, USA.}
\date{\today}
\myfootnote{Email addresses: bajaj@cs.utexas.edu (C. Bajaj), and tw27479@utexas.edu (T. Wang).}

\maketitle

\begin{abstract}
    Fusing a low-resolution hyperspectral image (HSI) and a high-resolution multispectral image (MSI) of the same scene leads to a super-resolution image (SRI), which is information rich spatially and spectrally. In this paper, we super-resolve the HSI using the graph Laplacian defined on the MSI. Unlike many existing works, we don't assume prior knowledge about the spatial degradation from SRI to HSI, nor a perfectly aligned HSI and MSI pair. Our algorithm progressively alternates between finding the blur kernel and fusing HSI with MSI, generating accurate estimations of the blur kernel and the SRI at convergence. Experiments on various datasets demonstrate the advantages of the proposed algorithm in the quality of fusion and its capability in dealing with unknown spatial degradation.
\end{abstract}

\section{Introduction}

\subsection{Hyperspectral-Multispectral Image Fusion}

Hyperspectral-multispectral image fusion \cite{yokoya2017hyperspectral} has been actively investigated in the remote sensing field. The goal is to fuse a low spatial resolution hyperspectral image (HSI) $\mathcal{Y}\in\mathbb{R}^{n_1\times n_2\times N_3}$ with a high spatial resolution multispectral image (MSI) $\mathcal{Z}\in\mathbb{R}^{N_1\times N_2\times n_3}$ of the same scene, to generate a spatio-spectral super-resolution image (SRI) $\mathcal{X}\in\mathbb{R}^{N_1\times N_2\times N_3}$. 

The problem is closely related to hyperspectral pan-sharpening \cite{loncan2015hyperspectral}, where a panchromatic image is utilized to enhance the spatial resolution of the HSI. Pan-sharpening methods such as \cite{aiazzi2006mtf,aiazzi2007improving,liu2000smoothing} can be extended to the multispectral case. 

Another type of approach utilizes the underlying low rank structure of the matricized SRI. Under the linear mixture model, every pixel vector (fiber in the spectral mode) is a convex combination of the spectral signatures of the underlying materials. Thus the matricized SRI, unfolded in the spectral mode, is of low rank. This representation originates from hyperspectral unmixing \cite{bioucas2012hyperspectral}, and has been adopted by, e.g., \cite{lanaras2015hyperspectral,HySure,yokoya2012coupled} to solve the fusion problem. 

In the review article \cite{yokoya2017hyperspectral}, the method \cite{HySure} is reported to show the most consistent and high performances in all the tests including visual, statistical, and classification-based assessment, amongst its competitors \cite{aiazzi2006mtf,aiazzi2007improving,akhtar2014sparse,eismann2003resolution,lanaras2015hyperspectral,liu2000smoothing,wei2015fast,yokoya2012coupled}. 

Recent efforts are directed to  working with HSI and MSI as tensors, and utilizing tensor decompositions \cite{kolda2009tensor} in either CP (CANDECOMP/PARAFAC) or Tucker formats. Dian et al. present a tensor sparse coding algorithm \cite{Dian2017} based on the Tucker decomposition. Later \cite{Dian2017} was further improved by \cite{Dian2018}, which minimizes a coupled objective function promoting sparsity in the core tensor of the Tucker decomposition. In the paper \cite{STEREO}, Kanatsoulis et al. utilizes the CP decomposition of the SRI. 

Many existing papers assume prior information about the spatial degradation from SRI to HSI, and/or use a perfectly aligned HSI and MSI pair. In practice, HSI and MSI are usually aligned only to some extent. Hence, further steps are required to co-register the images precisely. Moreover, it may not be practical to have a good description of the point spread function of the imaging system. Our main focus of this paper is thus to provide an approach that can perform fusion without prior knowledge of the spatial degradation nor the spectral degradation (band synthesis).

\subsection{Related Works and Our Main Contributions}

To our knowledge, dTV \cite{bungert2018blind}, STEREO \cite{STEREO} and HySure \cite{HySure} can deal with some unknown spatial degradation. In the literature, spatial degradation from SRI to HSI is usually modeled by the convolution of every band of the SRI using a blur kernel (small matrix, nonnegative, sum one), followed by downsampling. We assume the SRI is always spatially aligned with the MSI. If the blur kernel is not restricted to be centered at origin, it can compensate some translation error between the HSI and the SRI (and therefore also the MSI). The blur kernel is called separable if it can be decomposed into the inner product of two vectors. Gaussian kernels, centered at origin or not, are separable. STEREO assumes the blur is separable, which may not be the case in practice. dTV and HySure do not make such separability assumptions, thus they can handle broader types of blurs.

The spectral degradation from the SRI to the MSI can be modeled by a weighted summation of the hyperspectral bands according to the spectral responses of the multispectral sensor. STEREO assumes the availability of such information. HySure provides a way to estimate the spectral response. However, it still needs the spectral coverage information of the multispectral bands. So does dTV, which uses the directional total variation \cite{ehrhardt2016multicontrast} defined on each of the multispectral bands to super-resolve the bands of the HSI within the spectral coverage. 

Our proposed approach searches for the blur kernel and uses the graph Laplacian defined on the MSI to guide the super-resolution of all the bands of the HSI simultaneously. The proposed approach is able to achieve super-resolution without prior knowledge of the spatial nor spectral degradation. Table~\ref{tab:comparison} summarizes the capabilities of the most relevant prior work, and our method, and in dealing with unknown spatial and spectral degradation.   

\begin{table}[htp]
\caption{Summary of the types of unknowns different methods can deal with.}
\label{tab:comparison}
\begin{center}
\begin{tabular}{| c | c | c | c |}
\hline 
& Unknown Spatial Blur & Translation Error & Unknown Spectral Responses \\
\hline
dTV    & \cmark & \cmark & Spectral Coverage \\
\hline
STEREO & \text{Separable} & \cmark & \xmark \\
\hline
HySure & \cmark & \cmark & Spectral Coverage \\
\hline
\textbf{Ours} & \cmark & \cmark & \cmark \\
\hline
\end{tabular}
\end{center}
\end{table}

We explain the graph Laplacian regularization in Section 2, and present our super-resolution algorithm in Section 3. Empirical evaluations of our algorithm are conducted for a variety of data sets in Section 4. Conclusions are stated in Section 5. 

\section{Graph Laplacian Regularization}

The graph Laplacian is well known for its usefulness in spectral clustering \cite{von2007tutorial}, among many other applications. In the remote sensing field, it has been used by \cite{liao2018constrained} to convert a hyperspectral image to RGB for better visualization. Assuming the unknown SRI is aligned spatially with the MSI, we exploit the correlation between the SRI and MSI using the graph Laplacian.

\begin{figure}[htp]
\begin{center}
    \includegraphics[width=1\linewidth]{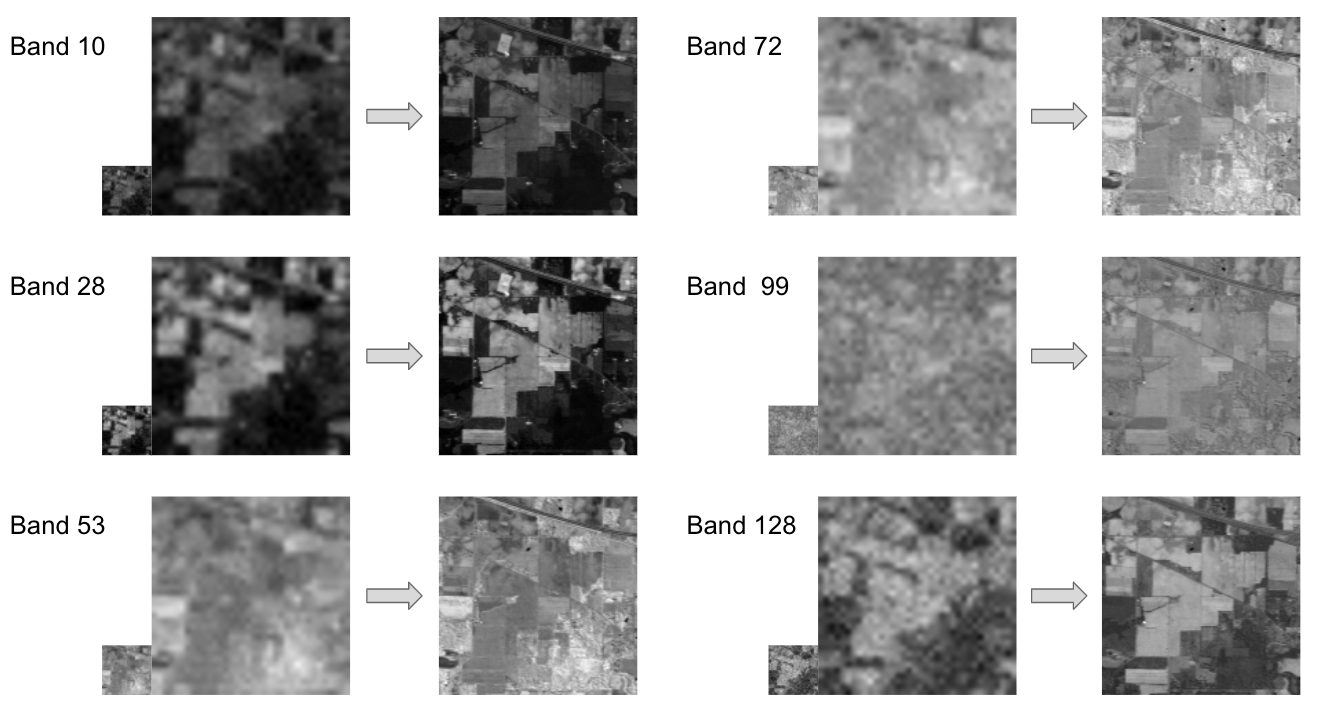}
\end{center}
\caption{Display of selected bands of the reconstructed SRI by solving \eqref{obj:non-blind}, on the Indian Pines dataset.
}
\label{fig:demonstration}
\end{figure}

Let $\bm{Y}\in\mathbb{R}^{(n_1n_2)\times N_3}$, $\bm{Z}\in\mathbb{R}^{(N_1N_2)\times n_3}$, and $\bm{X}\in\mathbb{R}^{(N_1N_2)\times N_3}$ be the matricization of $\mathcal{Y}$, $\mathcal{Z}$, and $\mathcal{X}$, respectively. Denote $\mathcal{L}(\bm{Z})\in\mathbb{R}^{(N_1N_2)\times(N_1N_2)}$ to be the graph Laplacian defined on the pixel vectors (i.e., rows) of $\bm{Z}$. There are many suitable choices for $\mathcal{L}(\bm{Z})$. Readers are referred to Section~\ref{sec:details} for our choice of the definition. Suppose $\bm{W}$ is the corresponding adjacency matrix of $\mathcal{L}(\bm{Z})$. The entry at $i$-th row and $j$-th column of $\bm{W}$, denoted by $w_{ij}$, encodes the similarity between the pixels vectors at these two spatial locations. We then apply the following graph Laplacian regularization to the SRI,
\begin{align*}
    \text{Tr}(\bm{X}^T\mathcal{L}(\bm{Z})\bm{X}) = \langle\bm{X},\mathcal{L}(\bm{Z})\bm{X} \rangle = \sum_{l=1}^{N_3}\left(\frac{1}{2}\sum_{i,j} w_{ij} (\bm{X}^{(i,l)}-\bm{X}^{(j,l)})^2\right)=\frac{1}{2}\sum_{i,j} w_{ij} \|\bm{X}^{(i,:)}-\bm{X}^{(j,:)}\|_2^2
\end{align*}
where $\bm{X}^{(i,:)}$ and $\bm{X}^{(j,:)}$ are the pixel vectors of $\bm{X}$ at $i$-th and $j$-th locations, respectively. Minimizing with respect to such regularization will force the pixel vectors that are close in the MSI to be close in the SRI, since the corresponding weights are larger.

For proof of concept, let us suppose the spatial degradation is known and the corresponding blur matrix is $\bm{C}\in\mathbb{R}^{(N_1N_2)\times(N_1N_2)}$. Denote $\bm{P}\in\mathbb{R}^{(n_1n_2)\times(N_1N_2)}$ to be the downsampling matrix. Then we can recover $\bm{X}$ by directly solving
\begin{align}\label{obj:non-blind}
    \min_{\bm{X}}~\|(\bm{P}\bm{C})\bm{X}-\bm{Y}\|_F^2+\alpha\text{Tr}(\bm{X}^T\mathcal{L}(\bm{Z})\bm{X}).
\end{align} 
Experiments with a simulated HSI and MSI pair, generated from the Indian Pines dataset, show that all bands are super-resolved simultaneously, via our graph Laplacian regularization. The results on selected bands are shown in Figure~\ref{fig:demonstration}.

\section{Blind Graph Laplacian Regularized Fusion (BGLRF)}

\subsection{Formulation}

When the spatial degradation is known, we solve \eqref{obj:non-blind} to reconstruct $\bm{X}$. When the spatial degradation is unknown, we alternate between finding the spatial blur and performing super-resolution. 

In the sequel, we will assume periodic boundary condition since it allows for convolution using Fast Fourier Transform (FFT). One can also modify the objective function such that it only penalizes at the region unaffected by the assumed boundary condition. Suppose $\bm{K}$ is the blur kernel, its convolution with the $l$-th band of the SRI can be written as
$$
\mathcal{C}(\bm{K})\cdot\bm{X}^{(:,l)},
$$
where $\mathcal{C}(\bm{K})\in\mathbb{R}^{(N_1N_2)\times(N_1N_2)}$ is a block circulant block matrix (BCCB) formed by the blur kernel $\bm{K}$, and $\bm{X}^{(:,l)}\in\mathbb{R}^{(N_1N_2)\times 1}$ is the vectorization of the SRI at the $l$-th band. In practice, we do not need to explicitly form the BCCB matrix. Convolution can be done by taking the inverse FFT of the entry-wise multiplication between the FFT of the zero-padded and circularly shifted blur kernel, and the FFT of the image. See Chapter 4 of \cite{hansen2006deblurring} for detailed explanations. 
With an estimate of the blur kernel size, denoted by $p$, we search for the best one among all the blur kernels of size no more than $p$. Similar to \cite{bungert2018blind,HySure}, we use isotropic total variation (TV) regularization for $\bm{K}$. TV regularization helps ensure the piece-wise smoothness of $\bm{K}$. We further restrain $\bm{K}$ to be in the simplex $\mathbb{S} \subset\mathbb{R}^{p\times p}$, where
$$
\mathbb{S} = \left\{\bm{K}\in\mathbb{R}^{p\times p}~|~\bm{K}^{(a,b)}\geq 0,~\sum_a\sum_b\bm{K}^{(a,b)} = 1\right\}.
$$
The simplex constraint ensures not only the physical meaning of the blur, but also the correct scale of the sought-after SRI. Denote $\text{I}_{\mathbb{S}}(\cdot)$ to be the indicator function of the simplex (0 inside the simplex, $\infty$ outside the simplex). When the spatial blur is unknown, we solve
\begin{equation}
\label{obj:blind}
\begin{aligned}
    \min_{\bm{K},\bm{X}}&~\|\bm{P}\mathcal{C}(\bm{K})\bm{X}-\bm{Y}\|_F^2+\alpha\text{Tr}(\bm{X}^T\mathcal{L}(\bm{Z})\bm{X})+\beta\text{TV}(\bm{K})+\text{I}_{\mathbb{S}}(\bm{K})
\end{aligned}
\end{equation}
to estimate both $\bm{K}$ and $\bm{X}$, where $\alpha>0$ and $\beta>0$ are some proper constants. 

\subsection{Solving \eqref{obj:blind}}

We solve \eqref{obj:blind} by proximal alternating minimization \cite{attouch2010proximal}. After initializing $\bm{X}$, we alternatively update $\bm{K}$ and $\bm{X}$ until convergence.

\subsubsection{Initialization} 
We initialize each band of $\bm{X}$ to be the bicubic interpolation of the corresponding band of $\bm{Y}$. 

\subsubsection{Update $\bm{K}$} 
Let $\bm{K}_{pre}$ be the previous estimate of $\bm{K}$. With an inertia term added at the end, we solve
\begin{align}
\label{update_K_original}
    \min_{\bm{K}}&~\|\bm{P}\mathcal{C}(\bm{K})\bm{X}-\bm{Y}\|_F^2+\alpha\text{Tr}(\bm{X}^T\mathcal{L}(\bm{Z})\bm{X})+\beta\text{TV}(\bm{K})+\text{I}_{\mathbb{S}}(\bm{K})+\tau\|\bm{K}-\bm{K}_{pre}\|_F^2
\end{align}
for some $\tau>0$. Denote $\mathcal{J}$ to be the operator that zero pad $\bm{K}$ and circularly shift it according to the center of the image of size $N_1\times N_2$. With a slight abuse of notation,
\begin{align*}
\{\mathcal{C}(\bm{K})\bm{X}\}^{(:,l)} = \mathcal{C}(\bm{X}^{(:,l)})\mathcal{J}(\bm{K})
\end{align*}
due to the property of circular convolution. Thus \eqref{update_K_original} is equivalent to
\begin{equation}
\label{update_K}
\begin{aligned}
    \min_{\bm{K}}&~\sum_{l}\|\bm{P}\mathcal{C}(\bm{X}^{(:,l)})\mathcal{J}(\bm{K})-\bm{Y}^{(:,l)}\|_F^2+\beta\text{TV}(\bm{K})+\text{I}_{\mathbb{S}}(\bm{K})+\tau\|\bm{K}-\bm{K}_{pre}\|_F^2,
\end{aligned}
\end{equation}
which is a non-smooth convex optimization, and can be solved via ADMM \cite{boyd2011distributed}. Define $\bm{G}=\mathcal{D}(\bm{K})\in\mathbb{R}^{(p^2)\times 2}$, where $\mathcal{D}$ is the operator that calculates the horizontal and vertical differences. Also introduce $\underline{\bm{K}}$ for the simplex constraint. The Lagrangian is equal to
\begin{align*}
    &\sum_{l}\|\bm{P}\mathcal{C}(\bm{X}^{(:,l)})\mathcal{J}(\bm{K})-\bm{Y}^{(:,l)}\|_F^2+\tau\|\bm{K}-\bm{K}_{pre}\|_F^2\\
    &+\beta\|\bm{G}\|_{2,1}+\mu\|\bm{G}+\bm{\Lambda}_1-\mathcal{D}(\bm{K})\|_F^2+\text{I}_{\mathbb{S}}(\underline{\bm{K}})+\mu\|\underline{\bm{K}}+\bm{\Lambda}_2-\bm{K}\|_F^2,
\end{align*}
where $\|\cdot\|_{2,1}$ is the summation of the 2 norm of the rows.
\begin{itemize}
    \item  Denote the adjoint operator of $\mathcal{J}$ to be $\mathcal{J}^*$. We find the $\bm{K}$ satisfying
    \begin{equation}
    \label{CG_K}
    \begin{aligned}
        &\sum_l\mathcal{J}^{*}\{(\mathcal{C}(\bm{X}^{(:,l)}))^*\bm{P}^{*}\bm{P}\mathcal{C}(\bm{X}^{(:,l)})\mathcal{J}(\bm{K})\}+\mu\mathcal{D}^{*}\mathcal{D}(\bm{K})+(\tau+\mu)\bm{K}\\
        =&\sum_l\mathcal{J}^{*}\{(\mathcal{C}(\bm{X}^{(:,l)}))^*\bm{P}^{*}\bm{Y}^{(:,l)}\}+\mu\mathcal{D}^*(\bm{G}+\bm{\Lambda}_1)+\mu(\underline{\bm{K}}+\bm{\Lambda}_2)+\tau \bm{K}_{pre}
    \end{aligned}
    \end{equation}
    using the conjugate gradient (CG) method. The multiplication with $\mathcal{C}(\bm{X}^{(:,l)})$ and its conjugate transpose $(\mathcal{C}(\bm{X}^{(:,l)}))^*$ can be done via FFT. Note that although \eqref{CG_K} is about solving for $\bm{K}$ of size $p\times p$ (small compared to $N_1$, $N_2$), one iteration of CG requires $O(N_1N_2N_3\log(N_1N_2))$ flops. 
    \item $\bm{G}$ can be updated by
    \begin{align*}
        \text{soft}(\mathcal{D}(\bm{K})-\bm{\Lambda}_1,\beta/2\mu),
    \end{align*}
    where $\text{soft}(\cdot,t)$, when applied to the gradient $(d_x,d_y)$ at position $(x,y)$, is equal to 
    $$\max\left(1-t/\sqrt{d_x^2+d_y^2},0\right)\cdot(d_x,d_y).$$
    \item $\underline{\bm{K}}$ can be updated by projecting
    $(\bm{K}-\bm{\Lambda}_2)$ into the simplex $\mathbb{S}$. We use the algorithm based on sorting, as described in \cite{wang2013projection}.
    \item To update the multipliers,
    \begin{align*}
        \bm{\Lambda}_1 \leftarrow& \bm{\Lambda}_1+\bm{G}-\mathcal{D}(\bm{K}),\\
        \bm{\Lambda}_2 \leftarrow& \bm{\Lambda}_2+\underline{\bm{K}}-\bm{K}.
    \end{align*}
\end{itemize}

\subsubsection{Update $\bm{X}$} 
Let $\bm{X}_{pre}$ be the previous estimate of $\bm{X}$. With an inertia term added at the end, we solve
\begin{align}
\label{update_X_original}
    \min_{\bm{X}}&~\|\bm{P}\mathcal{C}(\bm{K})\bm{X}-\bm{Y}\|_F^2+\alpha\text{Tr}(\bm{X}^T\mathcal{L}(\bm{Z})\bm{X})+\beta\text{TV}(\bm{K})+\text{I}_{\mathbb{S}}(\bm{K})+\tau\|\bm{X}-\bm{X}_{pre}\|_F^2,
\end{align}
for some $\tau>0$. This is equivalent to
\begin{equation}
\label{update_X}
\begin{aligned}
    \min_{\bm{X}}&~\|\bm{P}\mathcal{C}(\bm{K})\bm{X}-\bm{Y}\|_F^2+\alpha\text{Tr}(\bm{X}^T\mathcal{L}(\bm{Z})\bm{X})+\tau\|\bm{X}-\bm{X}_{pre}\|_F^2,
\end{aligned}
\end{equation}
which can by solved by finding the unique $\bm{X}$ satisfying
\begin{align}
\label{CG_X}
    (\mathcal{C}(\bm{K}))^*\bm{P}^{*}\bm{P}\mathcal{C}(\bm{K})\bm{X}+(\alpha \mathcal{L}(\bm{Z})+\tau\bm{I})\bm{X}
    = (\mathcal{C}(\bm{K}))^*\bm{P}^{*}\bm{Y}+\tau\bm{X}_{pre}.
\end{align}
We also use CG to solve \eqref{CG_X}. 

\subsection{Convergence}

The entire algorithm is summarized as follows. 

\begin{algorithm}[htp]
\caption{Blind Graph Laplacian Regularized Fusion (BGLRF)}
\label{alg:GLRF}
\begin{algorithmic}
\Statex \textbf{Initialize} $\bm{X}$ using bicubic interpolation
\For{$\text{iteration} = 0,1,\cdots$}\\
1. Update $\bm{K}$ by solving \eqref{update_K} using ADMM  \\
2. Update $\bm{X}$ by solving \eqref{update_X} using CG
\EndFor
\end{algorithmic}
\end{algorithm}

The objective in \eqref{obj:blind} is bi-convex. According to Proposition~1 in \cite{huang2016flexible}, we have the following result about the convergence of Algorithm~\ref{alg:GLRF}.

\begin{lemma}
\label{lem:convergence}
If the generated sequence $(\bm{K}$,$\bm{X})$ is bounded, Algorithm~\ref{alg:GLRF} converges to a stationary point of the objective function \eqref{obj:blind}.
\end{lemma}

\section{Numerical Experiments}

The experiments are executed from MATLAB R2018a on a 64-bit Linux machine with 8 Intel i7-7700 CPUs at 3.60 GHz and 32 GB of RAM.

\subsection{Simulated and Real Datasets}

We conduct experiments with simulated HSI and MSI pairs generated from the following data.

\textbf{Indian Pines} This hyperspectral image was acquired in 1992 by the AVRIS sensor over the Indian Pines test site 3 in northwestern Indiana \cite{PURR1947}. The sensor acquires 220 spectral bands, ranging from 400 to 2500 nm in wavelength. The ground sampling distance is 20 m, and the image is of size $145\times 145$. 200 bands are selected after removing bands of strong water vapor absorption. There are in total 16 classes of materials presented in the scene. 

\textbf{Salinas} This hyperspectral image was acquired by the AVIRIS sensor over Salinas Valley, California. There are 224 bands, covering 400 to 2500 nm. The spatial resolution is 3.7 m per pixel. The image is of size $512\times 217$. Similar as Indian Pines, 20 water absorption bands are discarded. There are also 16 classes in the scene.

\textbf{Pavia University} This hyperspectral image was captured over the University of Pavia, Italy, by the ROSIS-03 airborne instrument. The ROSIS-03 sensor has 115 channels with a spectral coverage ranging from 430 to 860 nm. 12 bands have been removed. The spatial resolution is 1.3 m per pixel. The spatial size is 640 $\times$ 340. There are 9 classes in total.

\begin{figure}[htp]
\begin{center}
    \includegraphics[width=1\linewidth]{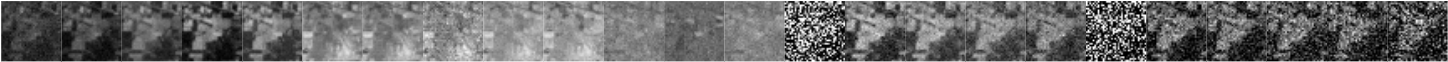}\\
   \includegraphics[width=1\linewidth]{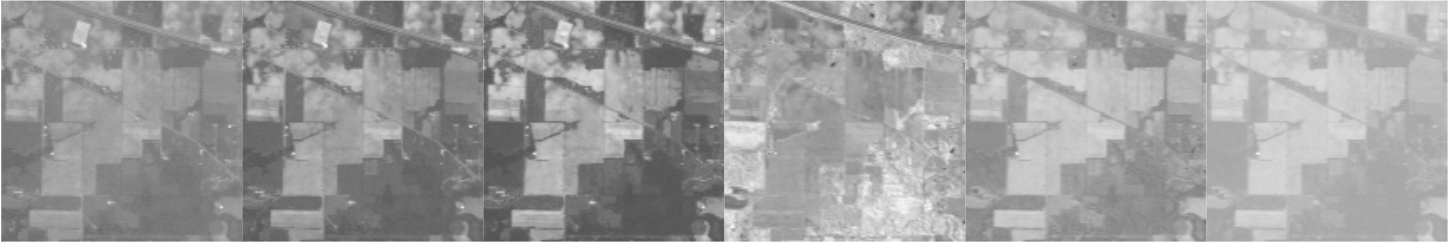}
\end{center}
\caption{(Top) Selected bands (every 8 band, starting from band 2) of the generated HSI from Indian Pines dataset; (Bottom) Generated MSI of 6 bands from Indian Pines dataset.}
\label{fig:IP_dataset}
\end{figure}

\begin{figure}[htp]
\begin{center}
    \includegraphics[width=1\linewidth]{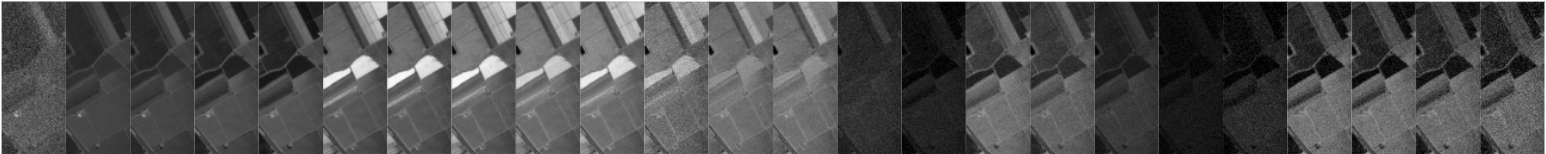}\\
   \includegraphics[width=1\linewidth]{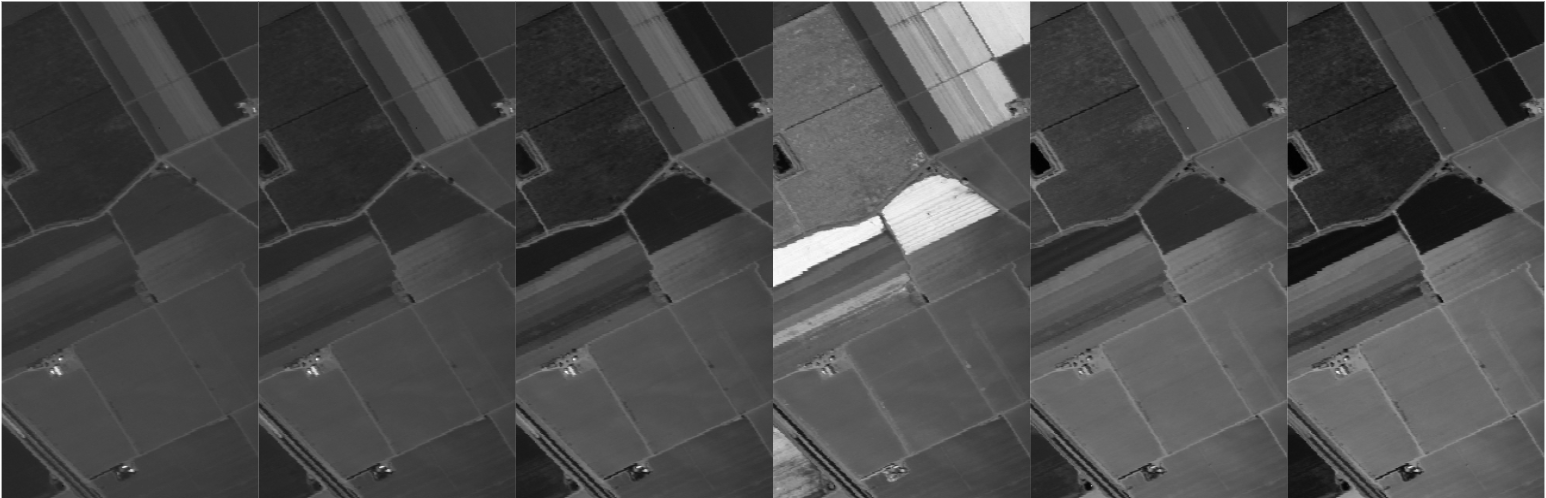}
\end{center}
\caption{(Top) Selected bands (every 8 band, starting from band 2) of the generated HSI from Salinas; (Bottom) Generated MSI of 6 bands from Salinas.}
\label{fig:Salinas_dataset}
\end{figure}

\begin{figure}[htp]
\begin{center}
    \includegraphics[width=1\linewidth]{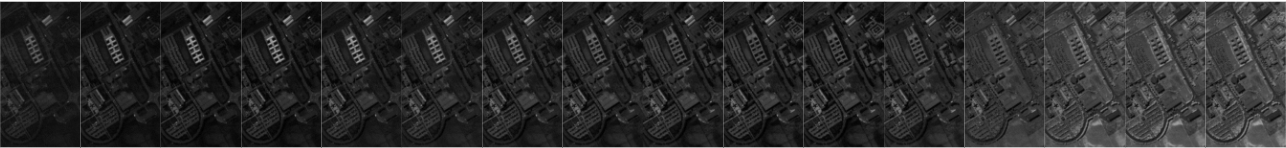}\\
   \includegraphics[width=1\linewidth]{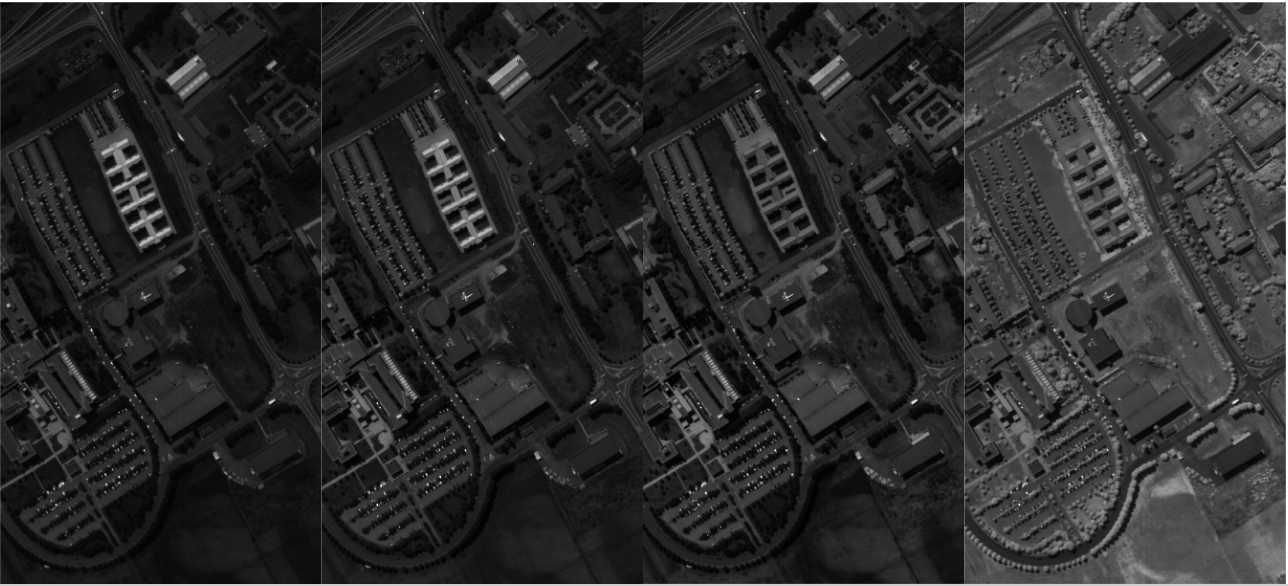}
\end{center}
\caption{(Top) Selected bands (every 6 band, starting from band 2) of the generated HSI from Pavia University dataset; (Bottom) Generated MSI of 4 bands from Pavia University dataset.}
\label{fig:PaviaU_dataset}
\end{figure}

We follow the standard protocol to generate simulated HSI and MSI pairs. Given the original hyperspectral image, the first step is to remove noise. We use the method \cite{roger1996principal} for its effectiveness and simplicity. HSI is the result of spatial degradation of an SRI. The degradation is usually modeled by some Gaussian blur, followed by downsampling plus additional noise. We define the standard deviation of the Gaussian blur such that its full width at half maximum (FWHM) is equal to the downsampling ratio $d$. The size of the Gaussian blur is assumed to be $(2d+1)$. $d$ is fixed to be 4 throughout the experiments. We then add Gaussian noise such that the HSI is of SNR 30. MSI is the synthesis of the spectral bands of SRI, plus additional noise. The synthesis is determined by the spectral responses of the multispectral sensor. We simulate such spectral responses by assuming the multispectral sensor is from some satellite. For the Indian Pines and Salinas, with spectral range from 400 to 2500 nm, we use the spectral responses from the Landsat 5 TM sensor. This sensor has spectral coverage approximately from 400 to 2400 nm, and generates a MSI of 6 bands (blue, green, red, near infrared, and two shortwave infrared bands). For the University of Pavia dataset, we use the spectral responses from the IKONOS sensor. This sensor has spectral coverage from 350 to 1100 nm, and generates a MSI of 4 bands (blue, green, red, near infrared). Gaussian noise is added such that the MSI is of SNR 40. Figure \ref{fig:IP_dataset}, \ref{fig:Salinas_dataset}, and \ref{fig:PaviaU_dataset} show the generated MSI and selected bands of the HSI of Indian Pines, Salinas, and Pavia University datasets, respectively.

We also test on real camera data. Since there was no paired HSI and MSI data available, we validate the performance of our proposed super-resolution algorithm on a  pair of multispectral (spectrally higher resolution) and panchromatic (spatially higher resolution) data.

\begin{figure}[htp]
\begin{center}
    \includegraphics[width=0.4\linewidth]{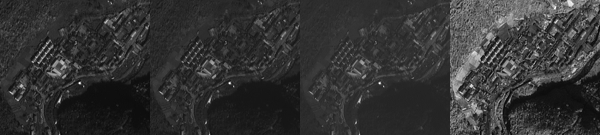}\\
   \includegraphics[width=0.4\linewidth]{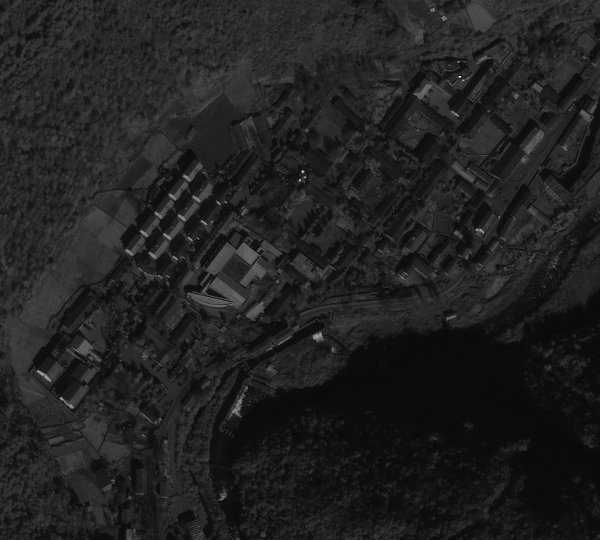}
\end{center}
\caption{(Top) The input blue, green, red and near infrared bands of the Western Sichuan dataset; (Bottom) The corresponding panchromatic input image of the Western Sichuan dataset.}
\label{fig:sichuan_dataset}
\end{figure}

\textbf{Western Sichuan} The data was acquired by an IKONOS-2 satellite over the western Sichuan area of China in 2008 \cite{IKONOS}. There are four multispectral bands, blue (455-520 nm), green (510-600 nm), red (630-700 nm), and near infrared (760-850 nm), with a ground sampling distance of 4 m. There is also one paired panchromatic image, with ground sampling distance of 1 m. Similar as \cite{chen2015sirf}, we use a selected region of size~$135\times 150$ from the multispectral image, and the corresponding region in the panchromatic image is of size~$540\times 600$. The four low resolution bands and the high resolution panchromatic image are shown in Figure~\ref{fig:sichuan_dataset}. There is some registration error between the two, allowing us to demonstrate the usefulness of our  blind super-resolution fusion algorithm.

\subsection{Implementation Details}\label{sec:details}

\noindent \textbf{Graph Laplacian} There are many suitable choices to generate the graph Laplacian on MSI. We choose to use the one presented in \cite{levin2008closed}\footnote{Code available at: http://www.alphamatting.com/code.php}, which defines the affinity of the pixel vectors using correlations within overlapping windows, utilizing both spectral and spatial information. To be specific, the $(i,j)$-th entry of $\mathcal{L}(\bm{Z})$ is defined using $\bm{p}_{i}=\bm{Z}^{(i,:)}$ and $\bm{p}_{j}=\bm{Z}^{(j,:)}$, and it's equal to
\begin{align*}
        \sum_{k|(i,j)\in \mathbb{W}_k}&\delta_{ij}-1/|\mathbb{W}_k|\cdot\Big(1+
    (\bm{p}_i-\bm{\mu}_k)\big(\bm{\Sigma}_k+\frac{\epsilon}{|\mathbb{W}_k|}\bm{I}\big)^{-1}(\bm{p}_j-\bm{\mu}_k)^T\Big),
\end{align*}
where the summation is with respect to all the overlapping windows $\mathbb{W}_k$ containing $\bm{p}_i$ and $\bm{p}_j$, $\delta_{ij}$ is 1 when $i=j$ and zero otherwise, $|\mathbb{W}_k|$ is the cardinality in the window, $\bm{u}_k$ is the mean in $\mathbb{W}_k$, $\bm{\Sigma}_k$ is the covariance, and $\epsilon$ is a small number to avoid numerical instability. The definition applies when $N_3$ is 1 (grayscale) or 3 (RGB), as considered in the original paper, and also for larger numbers when we have more bands. The resulting graph Laplacian is sparse, allowing for faster computation. The sparsity depends on the size of the overlapping windows.

\noindent \textbf{Conjugate Gradient} We use conjugate gradient (CG) to solve \eqref{CG_K} and \eqref{CG_X} for $\bm{K}$ and $\bm{X}$, respectively. We implement the CG solver ourselves. For $\bm{X}$ of very small size, it could be faster to explicitly generate the BCCB matrix $\mathcal{C}(\bm{K})$ and compute Choleksy decomposition, followed by forward/backward substitution to solve \eqref{CG_X}. 

\subsection{First Tests}

To better understand the necessity of blind kernel estimation and the graph Laplacian regularization, we generate a simulated HSI and MSI pair from the Indian Pines dataset, where the HSI is generated with a blur kernel shifted 4 pixels vertically and horizontally to the bottom right. We then consider the following two cases for comparisons.

\textbf{No GLR} We first consider the case where there is no graph Laplacian regularization (GLR). As a blind deconvolution problem, one solves
\begin{equation*}
\begin{aligned}
    \min_{\bm{X},\bm{K}}&~\|\bm{P}\mathcal{C}(\bm{K})\bm{X}-\bm{Y}\|_F^2+\beta\text{TV}(\bm{K})+\text{I}_{\mathbb{S}}(\bm{K}).
\end{aligned}
\end{equation*}
As mentioned in Section 3.1, TV regularization helps ensure the piece-wise smoothness of $\bm{K}$. The simplex constraint ($\bm{K}\in\mathbb{S}$ or $\text{I}_{\mathbb{S}}(\bm{K})=\infty$) ensures not only the physical meaning of the blur, but also the correct scale of the sought-after $\bm{X}$.

\textbf{Non-Blind} We also contrast our results with the case where there is no proper blur estimation. In this case, one solves  
\begin{align*}\label{obj:non-blind}
    \min_{\bm{X}}~\|\bm{P}\mathcal{C}(\bm{K})\bm{X}-\bm{Y}\|_F^2+\alpha\text{Tr}(\bm{X}^T\mathcal{L}(\bm{Z})\bm{X})
\end{align*}
with some pre-defined blur kernel $\bm{K}$. We assume the blur kernel centered at origin for $\bm{K}$, where the true blur kernel should be centered at bottom right instead. 

\begin{figure}[htp]
    \centering
    \subfigure[Bicubic Interpolation]{
    \includegraphics[width=0.24\linewidth]{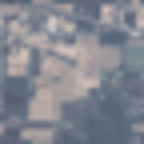}}    
    \subfigure[No GLR]{
    \includegraphics[width=0.24\linewidth]{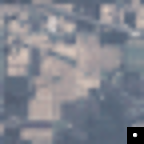}}    
    \subfigure[Non-Blind]{
    \includegraphics[width=0.24\linewidth]{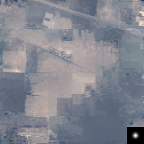}}    
    \subfigure[BGLRF]{
    \includegraphics[width=0.24\linewidth]{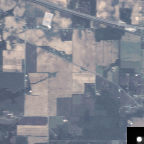}}    
\caption{The RGB bands generated from the spectral responses of the Landsat 5 TM sensor applied to: (a) Upsampled HSI via bicubic interpolation; (b) Super-resolution result without spatial information provided by the graph Laplacian regularization (along with the estimated blur kernel shown in right corner inset); (c) Super-resolution result using blur kernel incorrectly centered at origin;  (d) Super-resolution result by BGLRF (along with the estimated blur kernel shown in right corner inset).}
\label{fig:comparison}
\end{figure}

The comparisons are shown in Figure~\ref{fig:comparison}. To visualize the results, we apply the spectral responses of the multispectral sensor to the estimated SRIs. When there is no graph Laplacian regularization, the regularization about the blur kernel is not as useful. One cannot expect to get good estimation of the blur kernel due to the lack of spatial information, which is provided via the graph Laplacian. When there is no blur kernel estimation, the super-resolution is not as good.  

\subsection{Methods for Comparisons}

As mentioned in Section 1.2, dTV \cite{bungert2018blind}, STEREO \cite{STEREO} and HySure \cite{HySure} are the most related to our algoritm. 

dTV uses the directional total variation defined on each of the multispectral band to super-resolve the bands of the HSI within the same spectral range. The rationale is that the directional total variation carries location and direction information of the edges, which are similar for all the bands within the same spectral range. The fusion is performed band by band and we suspect that the spectral signatures may not be well-preserved in this manner, which could lead to suboptimal results in classification tasks. Also it is quite slow in our tests. Thus we choose not to compare with dTV.

STEREO assumes the availability of the spectral responses. When the blur kernel is known and separable, STEREO finds the factor matrices $\bm{A}$, $\bm{B}$ and $\bm{C}$ of the SRI tensor $\mathcal{X}$ by alternatively minimizing a coupled objective function with respect to the three factors. When the spatial degradation is unknown, STEREO needs to find two additional factors $\widetilde{\bm{A}}$ and $\widetilde{\bm{B}}$ corresponding to the HSI. The factors $\widetilde{\bm{A}}$ and $\widetilde{\bm{B}}$ are initialized by the CP decomposition of the MSI, followed by averaging according to the downsampling ratio. The codes are provided to us by the authors for the non-blind case. The CP tensor decomposition used for initialization is computed by TensorLab\footnote{Available at: https://www.tensorlab.net/}. The maximum iteration number is set to be 25. For successive iterations, the algorithm is stopped when the relative change in function value is less then $10^{-2}$. For the blind case, we implement the codes ourselves based on the description given in the original paper. The algorithm is stopped when the relative change in function value is less then $10^{-3}$. Through trail and error, we choose the CP rank that gives good quality metrics.

For HySure\footnote{Code available at: https://github.com/alfaiate/HySure}, we use the parameters suggested in the original paper. It first jointly estimates the spatial blur and the spectral responses, based on the relation that the spectrally degraded HSI is about the same as the spatially degraded MSI. HySure then seeks a piece-wise smooth abundance map of the spectral signatures of the SRI, by solving a coupled functional with TV regularization.

HySure has an implicit denoising capacity. As can be seen from Figure~\ref{fig:IP_dataset}, some bands of the HSI can be quite noisy. Thus we feed STEREO and our algorithm with the denoised HSI. The denoising is performed by \cite{roger1996principal}.

\subsection{Evaluation Metrics}

In the experiments with simulated data, where the true SRI is known, we use the following metrics based on spatial measures: \textbf{ERGAS} (relative dimensionless global error in synthesis \cite{wald2000quality}), \textbf{UIQI} (universal image quality index  \cite{wang2002universal}; We also use the following metrics based on spectral measures: \textbf{SAM} (spectral angle mapper \cite{kruse1993spectral}), \textbf{OA} (overall accuracy of classification). The SNR (signal-to-noise) ratio, measured in dB, is also adopted as a global error metric. Denote the estimated and ground truth SRI to be $\bm{X}$ and $\underline{\bm{X}}$, respectively.

\textbf{ERGAS} is defined as
\begin{align*}
100d\sqrt{\frac{1}{N_1N_2N_3}\sum\limits_{l=1}^{N_3}\frac{\|\bm{X}^{(:,l)}-\underline{\bm{X}}^{(:,l)}\|_F^2}{\mu_l^2}},
\end{align*}
where $d=N_1/n_1=N_2/n_2$ is the downsampling ratio, and $\mu_l$ is the mean in the $l$-th band. 

\textbf{UIQI} is the mean UIQI value of all bands, calculated between the estimated SRI and the ground truth. It takes values in $[-1,1]$, and measures spatial distortion between the two based on correlation, luminance and contrast. 

\textbf{SAM} measures the closeness (in degrees) of the estimated pixel vectors with the ground truth ones. The definition is
\begin{align*}
\frac{1}{N_1N_2}\sum\limits_{n=1}^{N_1N_2}\arccos\left(\frac{\langle\bm {X}^{(n,;)},\underline{\bm{X}}^{(n,:)}\rangle}{\|\bm{ X}^{(n,;)}\|_2\|\underline{\bm{X}}^{(n,:)}\|_2}\right).
\end{align*}

\textbf{OA} is the overall accuracy of classification. The benefits of fusion is in part bettering distinguishing different materials in the scene. While ground truth labels are available, the accuracy of fusion can be validated by the classification accuracy of the estimated SRI. We choose to use the binary SVM classifier and extend it to multi-class using the one-against-all (OAA) strategy \cite{melgani2004classification}. The SVM classifier is shown in \cite{ghamisi2017advanced} to still yield the state-of-art performances among the spectral classifiers. We train on the ground truth $\underline{\bm{X}}$ with $10\%$ randomly selected labels of each class (except for Indian Pines), and classify on the fused image $\bm{X}$ with the remaining labels. The value reported is averaged over 10 different sets of randomly chosen training samples. For Indian Pines, we discard classes 1, 4, 7, 9, 13, 15, 16 since their labels are relatively few.

\subsection{Results on Simulated and Real Data}

For simulated data, we compare the proposed algorithm with STEREO and HySure under different settings.

\begin{itemize}
    \item (mis-registration: 0) The first case is when the spatial degradation of HSI can be accurately estimated. Our algorithm is fed with the true blur kernel ($9\times 9$, centered at origin). We also feed STEREO and HySure with the ground truth blur kernel and spectral responses. 
    \item (mis-registration: 2) The center of the blur kernel is shifted horizontal and vertically to the top left by 2 pixels. We let HySure and our method search among all the blur kernels of size no more than 13. We feed STEREO and HySure with the ground truth spectral responses. 
    \item (mis-registration: 4) The center of the blur kernel is shifted horizontal and vertically to the bottom right by 4 pixels. We let HySure and our method search among all the blur kernels of size no more than 17. We feed STEREO and HySure with the ground truth spectral responses.
\end{itemize}

\begin{table}[htp]
\caption{Comparisons of the fusion results on Indian Pines, under different mis-registration values (shifts in blur kernel). ``GT'' stands for ground truth.}
\label{tab:IP}
\begin{center}
\begin{tabular}{| c | c | c | c | c | c | c | c |}
\hline
& \textbf{ERGAS} & \textbf{UIQI} & \textbf{SAM} & \textbf{OA~($\%$)} & \textbf{SNR~(dB)} & \textbf{TIME~(sec)} \\ 
\hline
GT & 0 & 1 & 0 & 85.14 & $\infty$ & - \\
\hline
\hline
& \multicolumn{6}{c|}{mis-registration: 0} \\
\hline
STEREO & 0.7334 & 0.8625 & 1.5712 & 63.55 & 30.6388 & \textbf{1.7262} \\
HySure & \textbf{0.5789} & \textbf{0.9097} & 1.2962 & 71.88 & 32.2836 & 8.6265 \\
\textbf{BGLRF} & 0.6053 & 0.8984 & \textbf{1.2686} & \textbf{75.20} & \textbf{32.4036} & 16.5606 \\
\hline
\hline
& \multicolumn{6}{c|}{mis-registration: 2} \\
\hline
STEREO & 0.8175 & 0.8459 & 1.7511 & 57.08 & 29.6748 & \textbf{3.0003} \\
HySure & 0.9318 & 0.8528 & 2.1029 & - & 27.0067 & 8.8469 \\
\textbf{BGLRF} & \textbf{0.6157} & \textbf{0.8988} & \textbf{1.2865} & \textbf{74.19} & \textbf{32.2442} & 192.267 \\
\hline
\hline
& \multicolumn{6}{c|}{mis-registration: 4} \\
\hline
STEREO & 0.9149 & 0.8181 & 1.9711 & 50.68 & 28.4043 & \textbf{2.8093} \\
HySure & 1.2792 & 0.7936 & 2.8644 & - & 24.0118 & 8.4512 \\
\textbf{BGLRF} & \textbf{0.6395} & \textbf{0.8969} & \textbf{1.3221} & \textbf{74.43} & \textbf{31.8393} & 310.996 \\
\hline
\end{tabular}
\end{center}
\end{table}

Table~\ref{tab:IP} shows the comparisons on the Indian Pines dataset. The CP rank is set to be 150 for STEREO. For our algorithm, we use $\alpha=10$ for all the settings, and $\beta=10$ for the mis-registration cases. When there is no error in the assumed blur kernel, HySure seems to be the best in terms of spatial metrics. It has a low ERGAS value of 0.5789, and a high UIQI value of 0.9097. Our method closely tracks HySure in spatial metrics (0.6053 and 0.8984, respectively), and outperforms HySure in terms of spectral metrics (1.2686 vs. 1.2962 for SAM, 75.20 vs. 71.88 for OA). When there is mis-registration (shift in blur kernel), our algorithm is consistently better in all the metrics except runtime. Our algorithm is also quite robust to the amount of mis-registration, since the quality metrics do not degrade much when the mis-registration increases. HySure does not handle large mis-registration well, which is in accordance with the results reported in \cite{yokoya2017hyperspectral}. Since the classification function is non-linear and possibly non-smooth, the OA values are only meaningful when the estimated SRI is in a small neighborhood of the ground truth. We omit the OA values for HySure in all the mis-registration cases.

Table~\ref{tab:Salinas} shows the comparisons on the Salinas. The CP rank is set to be 450 for STEREO. For our algorithm, we use $\alpha=10$ for all the settings, and $\beta=10$ for the mis-registration cases. Our algorithm is better in all the metrics except runtime. The other observations are similar as with the Indian Pines. One may notice the considerable increase in the runtime of our algorithm, this is due to its intrinsic computational complexity, and the need to iteratively refine the estimates of $\bm{K}$ and $\bm{X}$. Given the superior fusion quality, we think the extra time and efforts are worthwhile. 

\begin{table}[htp]
\caption{Comparisons of the fusion results on Salinas, under different mis-registration values (shifts in blur kernel). ``GT'' stands for ground truth.}
\label{tab:Salinas}
\begin{center}
\begin{tabular}{| c | c | c | c | c | c | c | c |}
\hline
& \textbf{ERGAS} & \textbf{UIQI} & \textbf{SAM} & \textbf{OA~($\%$)} & \textbf{SNR~(dB)} & \textbf{TIME~(sec)} \\ 
\hline
GT & 0 & 1 & 0 & 94.52 & $\infty$ & - \\
\hline
\hline
& \multicolumn{6}{c|}{mis-registration: 0} \\
\hline
STEREO & 1.5624 & 0.9282 & 0.8883 & 81.90 & 34.1961 & \textbf{17.1167} \\
HySure & 1.5952 & 0.9615 & 0.6148 & 83.81 & 37.5845 & 49.0683 \\
\textbf{BGLRF} & \textbf{1.3544} & \textbf{0.9555} & \textbf{0.5096} & \textbf{90.26} & \textbf{38.5692} & 283.800 \\
\hline
\hline
& \multicolumn{6}{c|}{mis-registration: 2} \\
\hline
STEREO & 1.6459 & 0.9010 & 1.4939 & 73.74 & 29.7953 & \textbf{22.2108} \\
HySure & 1.8587 & 0.9401 & 1.4046 & - & 27.1582 & 49.9079 \\
\textbf{BGLRF} & \textbf{1.3781} & \textbf{0.9554} & \textbf{0.5261} & \textbf{90.48} & \textbf{37.7534} & 1586.75 \\
\hline
\hline
& \multicolumn{6}{c|}{mis-registration: 4} \\
\hline
STEREO & 2.7020 & 0.7854 & 3.3530 & 57.79 & 22.1438 & \textbf{22.4562} \\
HySure & 2.4710 & 0.9054 & 2.1829 & - & 22.6494 & 50.8945 \\
\textbf{BGLRF} & \textbf{1.4225} & \textbf{0.9552} & \textbf{0.5611} & \textbf{89.42} & \textbf{35.9782} & 3314.64 \\
\hline
\end{tabular}
\end{center}
\end{table}

Table~\ref{tab:PaviaU} shows the comparisons on the Pavia University. The CP rank is set to be 550 for STEREO. For our algorithm, $\alpha=1$ for all the settings, and $\beta=10$ for the mis-registration cases. The observations are similar as with Salinas. When the mis-registration is 4, our bicubic initialization of $\bm{X}$ actually fails to produce good results. This is because the objective function \eqref{obj:blind} is non-convex (although bi-convex), and we need good initialization of $\bm{X}$ (or $\bm{K}$) in order to converge to the set of good solutions. For this particular case, we actually initialize $\bm{K}$ first, to be the blur kernel centered at origin. This example shows the limit of our method and its dependence on initialization for hard blind fusion problems. 

\begin{table}[htp]
\caption{Comparisons of the fusion results on the  Pavia University dataset, under different mis-registration values (shifts in blur kernel). ``GT'' stands for ground truth.}
\label{tab:PaviaU}
\begin{center}
\begin{tabular}{| c | c | c | c | c | c | c | c |}
\hline
& \textbf{ERGAS} & \textbf{UIQI} & \textbf{SAM} & \textbf{OA~($\%$)} & \textbf{SNR~(dB)} & \textbf{TIME~(sec)} \\ 
\hline
GT & 0 & 1 & 0 & 94.26 & $\infty$ & - \\
\hline
\hline
& \multicolumn{6}{c|}{mis-registration: 0} \\
\hline
STEREO & 1.8805 & 0.9737 & 3.2800 & 79.17 & 25.6020 & \textbf{33.9035} \\
HySure & 1.5681 & 0.9818 & 2.5267 & 87.47 & 27.8160 & 70.5435 \\
\textbf{BGLRF} & \textbf{1.4484} & \textbf{0.9830} & \textbf{2.2932} & \textbf{94.25} & \textbf{28.5737} & 199.247 \\
\hline
\hline
& \multicolumn{6}{c|}{mis-registration: 2} \\
\hline
STEREO & 2.0792 & 0.9684 & 3.7846 & 63.93 & 24.2236 & \textbf{49.7736} \\
HySure & 3.9080 & 0.9262 & 6.2028 & - & 17.5816 & 59.2992 \\
\textbf{BGLRF} & \textbf{1.5554} & \textbf{0.9822} & \textbf{2.3066} & \textbf{93.89} & \textbf{27.6705} & 2070.64 \\
\hline
\hline
& \multicolumn{6}{c|}{mis-registration: 4} \\
\hline
STEREO & 3.0886 & 0.9332 & 6.0683 & 52.80 & 19.7752 & \textbf{45.6595} \\
HySure & 5.0041 & 0.8889 & 7.8097 & - & 15.7662 & 58.7552 \\
\textbf{BGLRF} & \textbf{1.8236} & \textbf{0.9808} & \textbf{2.3141} & \textbf{93.71} & \textbf{25.7339} & 4753.38 \\
\hline
\end{tabular}
\end{center}
\end{table}

From the experiment results, we would like to emphasize our algorithm's ability to preserve classification accuracy, as validated by OA, along with the small SAM errors and overall closeness measured by SNR. The benefits of fusion is in part bettering distinguishing different materials in the scene. We consider our algorithm to be suitable for this purpose.

\begin{figure}[htp]
\begin{center}
    \includegraphics[width=0.16\linewidth]{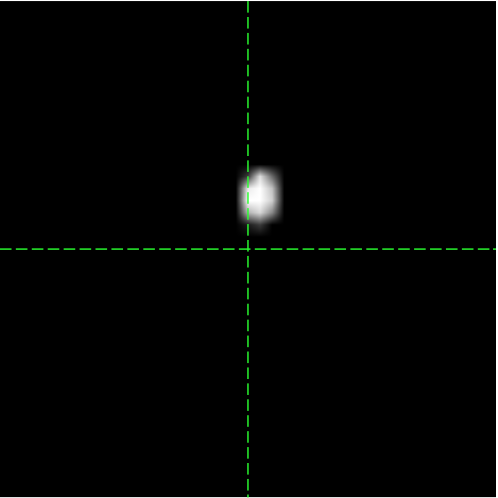}\\ 
    \includegraphics[width=0.4\linewidth]{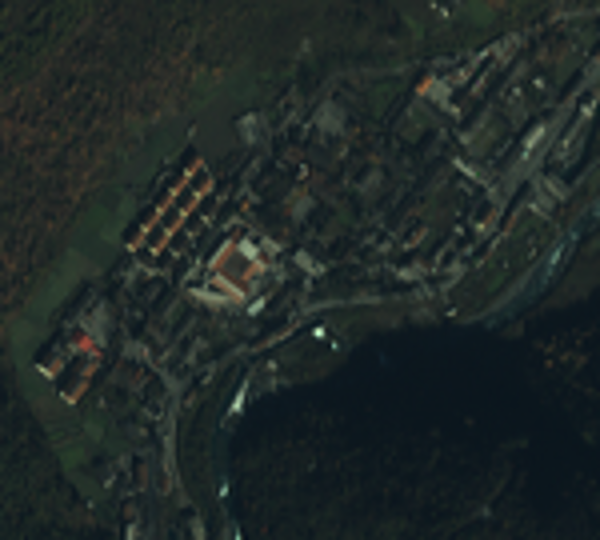}
    \includegraphics[width=0.4\linewidth]{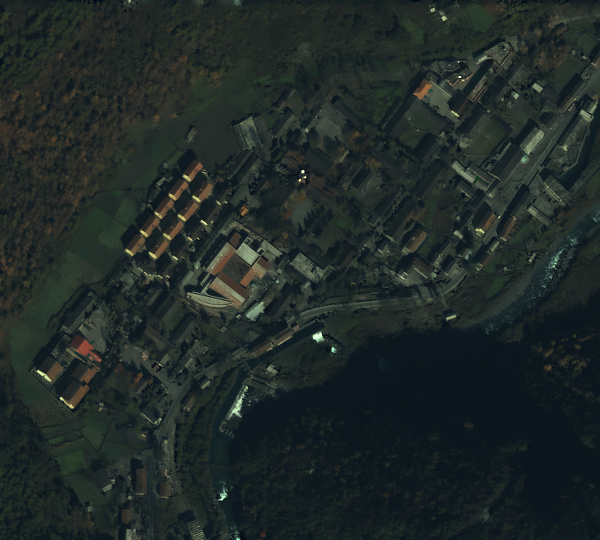}\\
    \includegraphics[width=0.4\linewidth]{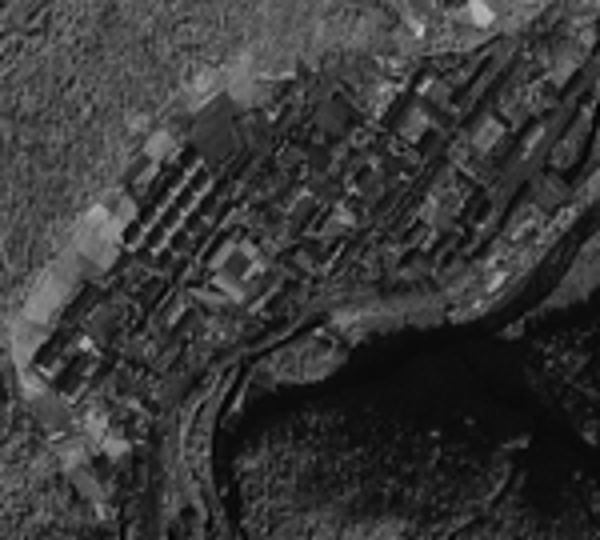}
    \includegraphics[width=0.4\linewidth]{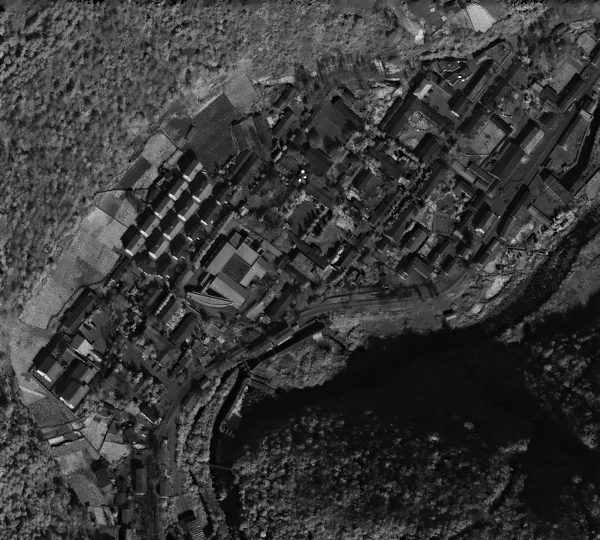}
\end{center}
\caption{Row 1: Estimated blur kernel of the Western Sichuan dataset (enlarged for better visualization); Row 2: Composition of the RGB bands from (Left) the bicubic interpolation of the input low-resolution bands and (Right) the super-resolved ones; Row 3: (Left) Bicubic interpolation of the input low-resolution near infrared band and (Right) the super-resolved near infrared band.}
\label{fig:sichuan_sr}
\end{figure}

For the real data, the fusion results on the blue, green, red and near infrared bands, along with the estimated kernel, are shown in Figure~\ref{fig:sichuan_sr}. We let our algorithm searches among all the blur kernels of size no more than 25. We can see that there is indeed some translation error between the MSI and panchromatic pair. This error is compensated by the proposed algorithm.

\section{Conclusion}

We presented a hyperspectral-multispectral fusion algorithm using graph Laplacian regularization, without assuming prior knowledge about the blur kernel. Our algorithm alternates between finding the blur kernel and fusing HSI with MSI. As a byproduct, our algorithm is able to deal with translation mis-alignment between the two input images. Various numerical experiments validate the usefulness of our algorithm, and its ability to preserve spectral information. Such property is desirable for further classification and detection tasks. 

\noindent \textbf{Acknowledgment} 
This research was supported in part by the grant from NIH - R01GM117594.

\newpage
\bibliographystyle{plain}
\bibliography{references}

\end{document}